\documentclass[manuscript,screen]{acmart}
\usepackage{multirow}
\setcopyright{acmcopyright}
\copyrightyear{2022}
\acmYear{2022}
\acmDOI{XXXXXXX.XXXXXXX}

\begin{document}

\title{Pose Uncertainty Aware Movement Synchrony Estimation via Spatial-Temporal Graph Transformer}
% \title{Pose Uncertainty-Aware Contrastive Movement Synchrony Estimation via Spatial-Temporal Graph Transformer}
% \title{Joint Uncertainty-Aware Contrastive Movement Synchrony Estimation via Spatial-Temporal Graph Transformer}

\author{Jicheng Li}
\email{lijichen@udel.edu}
\affiliation{%
  \institution{University of Delaware}
  \streetaddress{18 Amstel Ave}
  \city{Newark}
  \state{Delaware}
  \country{United States}
  \postcode{19716}
}

\author{Anjana Bhat}
\email{abhat@udel.edu}
\affiliation{%
  \institution{University of Delaware}
  \streetaddress{540 S. College Ave} %address of health sciences
  \city{Newark}
  \state{Delaware}
  \country{United States}
  \postcode{19713}
}

\author{Roghayeh Barmaki}
\email{rlb@udel.edu}
% Corresponding author
\authornote{Correspondence: Roghayeh Barmaki (rlb@udel.edu)}
\affiliation{%
  \institution{University of Delaware}
  \city{Newark}
  \state{Delaware}
  \country{United States}
}

\renewcommand{\shortauthors}{Li et al.}
\newcommand{\etal}{\textit{et al. }}

\begin{abstract}
Movement synchrony reflects the coordination of body movements between interacting dyads. 
The estimation of movement synchrony has been automated by powerful deep learning models such as transformer networks.
However, instead of designing a specialized network for movement synchrony estimation, previous transformer-based works broadly adopted architectures from other tasks such as human activity recognition.
% model: components
Therefore, this paper proposed a skeleton-based graph transformer for movement synchrony estimation. 
The proposed model applied ST-GCN, a spatial-temporal graph convolutional neural network for skeleton feature extraction, followed by a spatial transformer for spatial feature generation. 
The spatial transformer is guided by a uniquely designed joint position embedding shared between the same joints of interacting individuals.  
Besides, we incorporated a temporal similarity matrix in temporal attention computation considering the periodic intrinsic of body movements.
% uncertainty
In addition, the confidence score associated with each joint reflects the uncertainty of a pose, while previous works on movement synchrony estimation have not sufficiently emphasized this point.
We introduced a pose uncertainty awareness mechanism in spatial and temporal feature generations. 
% pretraining
Since transformer networks demand a significant amount of data to train, we constructed a dataset for movement synchrony estimation using Human3.6M, a benchmark dataset for human activity recognition, and pretrained our model on it using contrastive learning. 
% knowledge
We further applied knowledge distillation to alleviate information loss introduced by pose detector failure in a privacy-preserving way.
We compared our method with representative approaches on PT13, a dataset collected from autism therapy interventions. 
Our method achieved an overall accuracy of $88.98\%$ and surpassed its counterparts by a wide margin while maintaining data privacy. This work has implications on synchronous movement activity recognition in group settings, with applications in education and sports.
\end{abstract}

\begin{CCSXML}
<ccs2012>
   <concept>
       <concept_id>10003456.10010927.10003616</concept_id>
       <concept_desc>Social and professional topics~People with disabilities</concept_desc>
       <concept_significance>500</concept_significance>
       </concept>
   <concept>
       <concept_id>10003120</concept_id>
       <concept_desc>Human-centered computing</concept_desc>
       <concept_significance>500</concept_significance>
       </concept>
   <concept>
       <concept_id>10010147.10010178.10010224.10010225.10010228</concept_id>
       <concept_desc>Computing methodologies~Activity recognition and understanding</concept_desc>
       <concept_significance>500</concept_significance>
       </concept>
</ccs2012>
\end{CCSXML}

\ccsdesc[500]{Social and professional topics~People with disabilities}
\ccsdesc[500]{Human-centered computing}
\ccsdesc[500]{Computing methodologies~Activity recognition and understanding}

%%
%% Keywords.Separate the keywords with commas.
\keywords{deep learning, movement synchrony estimation, contrastive learning, transformer networks, knowledge distillation, autism spectrum disorder}

\maketitle

\section{Introduction}
% 1. concept of movement synchrony
Movement synchrony describes the coordination of body movements between interacting partners. 
In psychotherapeutic counseling, movement synchrony has been considered to reflect the depth of the client-counselor relationship \cite{nagaoka2008body}.
In sports, such as artistic swimming and synchronized diving, synchrony scores are often reported to reflect teamwork impressions.
The autism community also identifies movement synchrony as a key indicator of children's social and developmental achievements.
Specifically, this paper investigates movement synchrony between children and therapists in play therapy, a structured, theoretically based approach to therapy that builds upon children's regular communication and learning processes.
% 2. brief statistical method, deep learning method, transformer

% To automate the process of movement synchrony estimation, statistical approaches \cite{tarr2018synchrony, fujiwara2020rhythmic, altmann2021movement, chu2015unsupervised, sariyanidi2020discovering} have been used broadly. 
Statistical approaches \cite{tarr2018synchrony, fujiwara2020rhythmic, altmann2021movement, chu2015unsupervised, sariyanidi2020discovering} have been widely employed to automate movement synchrony estimation.
Such approaches convert intervention video recordings to pixel-wise representations to reflect temporal dynamics, and the final synchrony score is computed as the correlation between pixel sequences from different participants. Despite strengths, statistical approaches are significantly prone to noise since all the image pixels are considered and treated equally. 
This can be a severe problem for video recordings of a non-stationary camera, in which the background is dynamically changing. 
% On the other hand, approaches based on motion energy~\cite{tarr2018synchrony, fujiwara2020rhythmic, altmann2021movement} do require a fixed and pre-defined region of interest (ROI) and fail to work when participants move outside the ROI. 
Particularly, as one of the most popular statistical methods, motion energy based approaches ~\cite{tarr2018synchrony, fujiwara2020rhythmic, altmann2021movement} require a fixed and pre-defined area of interest (ROI) and fail to work when the participants move outside the ROI.
In addition, the topology relationship between different human body parts is neglected too. 

Recently, deep learning methods have gained popularity as a means to overcome the shortcomings of statistical approaches, as they have demonstrated superior performance in human activity comprehension tasks \cite{zheng2021poseformer,pavllo2019videopose3d,yan2018spatial,dwibedi2020counting}. 
Li \etal \cite{Li2021improving} introduced a multi-task learning approach for estimating movement synchrony from invention video recordings, and the follow-up work \cite{li2022dyadic} further addressed the privacy issue by only using secondary data (skeleton, optical flow) generated from original videos, given that raw video recordings may exposing one's appearance and identity. 
In particular, skeleton data, as a record of joint coordinates, has been extensively used in various human activity comprehension tasks \cite{yan2018spatial,plizzari2021skeletonSTTF, zheng2021poseformer,reily2018skeleton}.
It is also robust to changes in background and appearance \cite{zhang2017view} and provides a privacy-preserving, light-weight representation that can be processed faster than RGB videos \cite{thoker2021skeleton}. 
Therefore, in this work, we choose skeleton data as input for our model.

% 3. previous transformer method, our approach, update in transformer architecture:
% i. encode graph neural network to capture spatial topology
% ii. embed cross similarity matrix in temporal attention computation
% iii. pose uncertainty introduced by pose estimator
% 4. unique positional embedding
%spatial
Skeleton-based Transformer Networks (TFNs) \cite{vaswani2017attention} have shown state-of-the-art performance in multiple tasks such as pose estimation \cite{zheng2021poseformer} and activity recognition \cite{plizzari2021skeletonSTTF}.
Nonetheless, many studies \cite{zheng2021poseformer,plizzari2021skeletonSTTF,liu2022graph} only include a single skeleton from a single person as input, which needs to be adjusted to support movement synchrony estimation, a task involving two interacting partners, for example, a child and a therapist.
In addition, the topology relationship between joints has not been fully explored as well. 
% However, their work applied a simple linear projection layer to generate spatial embeddings from raw coordinates and did not consider topology relationship among joints. 
% ST-GCN
To address this issue, we adopted Spatial-temporal Graph Convolution Neural Network (ST-GCN) \cite{yan2018spatial} which takes human body’s natural connections into account to better capture joint topology.
%joint index embedding
We also designed a unique joint position embedding shared between the same joint of dyads to accommodate TFN to two people, as well as guiding our model to attend to the same joint in synchrony measurement.
% TSM in temporal attention 
Meanwhile, considering the periodic intrinsic of our dataset, which records periodic movements in autism therapy interventions such as squats and jumping, we incorporate a temporal similarity matrix (TSM) \cite{li2022dyadic,dwibedi2020counting,su2021how} in temporal attention computation to better capture temporal dynamics.
% TSM can help to regularize TFNs and guide temporal attention generation while offering the TFNs to better understand offset and resemblance of actions in temporal dimension, especially in periodic movements.
%pose uncertainty
Furthermore, prevailing pose estimators \cite{cao2019openpose, chen2018cpn, sun2019hrnet} typically associate a confidence score with each joint detection. 
The confidence score reflects the uncertainty of the predicted joint coordinates, while previous works on movement synchrony estimation \cite{li2022dyadic,Li2021improving,tarr2018synchrony, fujiwara2020rhythmic, altmann2021movement, chu2015unsupervised, sariyanidi2020discovering} have not sufficiently emphasized this aspect. 
This work integrates pose uncertainty both spatially and temporally: 
ST-GCN incorporates joint confidence into spatial feature generation, while a frame uncertainty embedding is used to represent overall confidence at a specific time stamp in temporal feature generation.
% ST-GCN incorporates joint confidence in spatial feature generation, and a frame uncertainty embedding is included in temporal feature generation to represent the overall confidence at a given time stamp. 

%4. TF pretraining, data augmentation
% Nevertheless, deep learning approaches are more resource-intensive in terms of data and computation capacity.
% Commonly used backbone networks \cite{Tran2015c3d,carreira2017i3d,vaswani2017attention,dosovitskiy2020vit} require a significant amount of labeled data (several hundreds of thousands, even millions) to achieve its best performance, which is not always feasible, for example, in autism research.
The downside of TFNs is that they are highly resource-intensive in data and computational resources.
Inadequately pretrained TFNs may not even produce competitive results to less complex counterparts such as Convolutional Neural Networks (CNNs) \cite{he2016ResNet}. 
%  One possible reason is that CNNs can encode prior knowledge to alleviate data consumption, while TFNs have to discover such information from significant large-scale dataset \cite{khan2021transformers}. 
 To solve this problem, a common practice is to first pretrain TFNs on a large corpus and subsequently fine-tune them to the dataset of interest, which is typically much smaller than the pretraining dataset \cite{khan2021transformers, liu2019roberta}. 
In this paper, we constructed data samples at different synchrony levels using Human3.6M \cite{Ionescu2014h36m}, a large-scale benchmark dataset for human activity recognition, and the new dataset is referred to as \textit{Human3.6MS} in this work. 
We first pretrained our model on Human3.6MS, and then fine-tuned it on PT13, a dataset on movement synchrony estimation in autism therapy. 
 %  Data construction is based on the assumption that each person in an interacting dyad represents a different view of the other person.
We further applied contrastive learning in pretraining, a machine learning technique to learn by comparing between different samples instead of learning from each data sample individually \cite{le2020contrastive}. 
Such comparison can be conducted between similar and dissimilar inputs, referred to as positive and negative pairs, respectively. 
With the aid of contrastive learning, our model can distinguish between data different levels of synchrony better and create a robust hidden embedding.

% 5. knowledge distillation
Lastly, one major limitation of skeleton data is information loss caused by pose detector failures, especially when the model encounters challenging, uncommon poses. 
Hence, in this paper, we suggest that a model trained on skeleton data can acquire missing knowledge from the one trained on original video recordings. 
We used knowledge distillation in this paper, a process of transferring knowledge from one  model (teacher) to another model (student) without compromising validity \cite{hinton2015distilling}. 
This approach has been widely used in model compression in the past \cite{Buciluundefined2006model,hinton2015distilling}. 
In this paper, we present the usage of knowledge distillation in a novel way to combat information loss arising from pose estimators while ensuring privacy-preserving neural network training.
We define a model trained on intervention video recordings (which is only accessible by therapists) as a \textit{"teacher"} model, and a model that only has access to privacy-preserving skeleton data as a \textit{"student"} model.
Information about a concise knowledge representation from the teacher model is encoded in the pseudolikelihoods assigned to its outputs, and such knowledge can be distilled into a student model by learning the predicted distribution (also called ``\textit{soft labels}") of the teacher model.

% 6. summarize contribution
In summary, our contributions are:
\begin{enumerate}
    \item We propose a skeleton-based graph transformer for movement synchrony estimation while addressing pose uncertainty. 
    \item We propose a pretraining strategy via contrastive learning using constructed data samples from Human3.6M, a large-scale benchmark dataset for human activity recognition.
    \item We present several synchrony-invariant methods for augmenting skeleton data, both spatially and temporally. 
    \item We apply knowledge distillation in a new way to mitigate the information loss arising from pose detector failures while preserving privacy. 
\end{enumerate}
    % \item We propose a pretraining strategy via contrastive learning using carefully computed data sample from publicly available benchmark datasets.
    % \item we introduce a way to leverage large-scale benchmarks to solve the data insufficiency problem in autism, especially for activity comprehension in therapy intervention.

% paper structure
The rest of the paper is organized as follows. Section \ref{sec:relatedwork} reviews related work. 
Section \ref{sec:method} introduces the proposed method.
% ,and section \ref{sec:dataset} describe the dataset. 
The experiment results are presented in Section \ref{sec:experiment}, followed by discussion and conclusion in section \ref{sec:discussion} and \ref{sec:conclusion}, respectively.

\section{Related Work}
\label{sec:relatedwork}
% mainly focus on deep learning methods
\subsection{Movement Synchrony Estimation}
Representing studies \cite{tarr2018synchrony, fujiwara2020rhythmic, altmann2021movement} widely applied statistical measurements such as correlation to quantify movement synchrony. Other works \cite{chu2015unsupervised, Chang_2019_CVPR, sariyanidi2020discovering} investigated synchrony from video via pixel-level knowledge. 
Since our model is based on deep neural networks, we particularly focus on movement synchrony methods using deep learning. 
Calabr{\`o} \etal \cite{calabro2021mms} investigated interaction between children and therapists by reconstructed electrocardiograms using a convolutional auto-encoder. 
Li \etal \cite{Li2021improving} proposed a multi-task framework to integrate movement synchrony estimation with two auxiliary tasks, activity recognition and action quality assessment. This multi-task paradigm further enhances the accuracy of movement synchrony estimation. They also applied distribution learning to mitigate human bias in synchrony data annotations by allowing multiple labels associated with a confidence score for each record. 
However, both works \cite{calabro2021mms,Li2021improving} require access to original video recordings, which can reveal one's identity and suffer from data sharing roadblock in autism research. To address the issue, a privacy-preserving ensemble method for movement synchrony estimation is proposed in \cite{li2022dyadic}, which relies entirely on publicly accessible, identity-agnostic secondary data such as optical flow and skeleton data. This work can also be generalized to athlete performance evaluation when synchrony scores are reported, for example, in synchronized diving. 
Park \etal \cite{park2021Similarity} explored a relevant task, human motion similarity, since movement synchrony estimation also implicitly involves measuring motion similarity in both spatial and temporal aspects. The proposed model in \cite{park2021Similarity} takes a sequence of human joints as input followed by body part decomposition, and then generates embeddings for each body part separately.
The motion embedding is produced by data reconstruction similar to \cite{calabro2021mms}, and a motion variation loss was introduced to further distinguish between similar motions.

\subsection{Skeleton-based Graph Transformer Networks}
% Skeleton-based human action recognition requires understanding relationships between different body joints within and across frames.
Yan \etal \cite{yan2018spatial} designed a generic representation of skeleton sequences for action recognition with a spatial-temporal graph model, and proposed ST-GCN that constructs a set of spatial temporal graph convolutions on the skeleton sequences. 
Despite its superior performance, limitations of ST-GCN are found in both spatial and temporal feature generation. For example, the graph representation of skeleton is fixed for all layers and actions while subject to biological connectivity, and convolution operations are operated in a local neighborhood depending on the kernel size.
Therefore, some recent work tried to overcome such limitations by introducing transformer networks. 
Plizzari \etal \cite{plizzari2021skeletonSTTF} extended Yan's work by replacing GCN with TFNs. The proposed model (ST-TR) includes a spatial self-attention module (SSA) to model relations between different joints, and a temporal self-attention module (TSA) to capture long-range inter-frame dependencies. Similar to \cite{yan2018spatial}, SSA models the correlation between body joints independently, while TSA focuses on the temporal dynamics of one particular joint. 
Liu \etal \cite{liu2022graph} proposed KA-AGTN which models spatial joint dependencies by an adaptive graph, and a graph transformer operator was applied to the graph for spatial representation generation, while temporal features were obtained by a channel-wise attention. 
Note that both ST-TR and KA-AGTN still extract low-level features through a standard ST-GCN.
Bai's work \cite{bai2021hierarchical} introduced a Hierarchical Graph Convolutional skeleton Transformer (HGCT) based on a disentangled spatiotemporal transformer (DSTT) block to overcome neighborhood constraints and entangled spatiotemporal feature representations of GCNs. Different from ST-GCN, the adjacency matrix of DSTT is parameterized and updated in the training process. STGC also allows flexible modeling of graph topology by fine-tuning local topology hierarchically. Besides, they also claim that HGCT is lightweight and computationally efficient.

\subsection{Contrastive Learning}
% A main purpose of unsupervised learning is to pretrain
% representations (i.e., features) that can be transferred to
% downstream tasks by fine-tuning.
Contrastive learning has been widely applied to unsupervised visual representation learning \cite{lai2019cpc,he2020moco,chen2020simclr,tian2020cmc}. In skeleton representation learning, Rao \etal \cite{rao2021Gait} proposed a self-supervised approach to learning effective gait representations. They explored several pretext tasks, including reverse skeleton sequence reconstruction, sorting, and prediction. A novel locality-aware attention
mechanism and locality-aware contrastive learning scheme are proposed to incorporate the intra-sequence and inter-sequence locality into gait encoding process.
In \cite{rao2022simmc} SimMC was presented to efficiently learn representations of unlabeled skeleton sequences for person reID. SimMC clusters the randomly masked skeleton sequences and contrasts their features with the most typical ones to learn discriminative skeleton representations leveraging inherent motion similarity and continuity. 
Thoker \etal \cite{thoker2021skeleton} proposed inter-skeleton contrastive learning, which learns from multiple different input skeleton representations in a cross-contrastive manner for action recognition. They also contributed several skeleton-specific spatial and temporal augmentations, including 3D shearing, joint jittering, and temporal cropping. 
Lin \etal \cite{lin2020ms2l} integrated contrastive learning with multi-task learning in action recognition. They selected two auxiliary tasks: motion prediction to model skeleton dynamics and jigsaw puzzles solving to capture temporal patterns.
Xu \etal \cite{xu2021prototypical} proposed PCRP which performs reverse motion prediction to extract discriminative temporal pattern. They also introduce action prototypes as latent variables and formulate PCRP training as an expectation-maximization (EM) task, where PCRP iteratively runs (1) E-step as to estimate the distribution of action encoding by K-means clustering, and (2) M-step as optimizing the model by minimizing the proposed ProtoMAE loss by contrastive learning. 

\subsection{Knowledge Distillation}
Existing works broadly adopted a ``teacher-student" paradigm, where each data owner learns a teacher model using its own (local) private dataset (such as medical records), and the data user aims to learn a student model to mimic the output of the teacher model(s) using the unlabelled public data. The student model can be publicly released instead.
Papernot's work \cite{papernot2017semisupervised} proposed a PATE approach, which combines multiple models trained with disjoint sensitive datasets in a black-box fashion. These models are not published  but instead are used as ``teachers” for a ``student” model. The student learns to predict an output chosen by noisy voting among all of the teachers, and cannot directly access an individual teacher or the underlying data or parameters.
Zhuang \etal \cite{zhuang2022locally} proposed a method called LDP-DL that also leveraged knowledge distillation similar to PATE. LDP-DL employed local differential privacy on the data owners' side without any trusted aggregator as required in PATE, and designed an active query sampling to reduce the total number of queries from data users to each data owner. 
% In Gao's work \cite{gao2020private} a private teacher trained on sensitive data is not publicly accessible but teach a student to be publicly released instead. 
Gao \etal \cite{gao2020private} provided a teacher-student-discriminator learning framework, where the student acquires the distilled knowledge from the teacher and is trained with the discriminator to generate similar outputs as the teacher. The framework allows a student to be trained with unlabelled public data in a few epochs.
Sui's work \cite{sui2020feded} proposed a privacy-preserving medical relation extraction model based on federated learning \cite{konevcny2016federated} which typically requires uploading local models' parameters for center model updates deployed on a server. To overcome the communication bottleneck caused by cumbersome uploads, they leveraged knowledge distillation to update the central model using predictions of ensemble local models instead of local parameters.

\section{Methods}
\label{sec:method}
\begin{figure}[h]
  \centering
  \includegraphics[width=\linewidth]{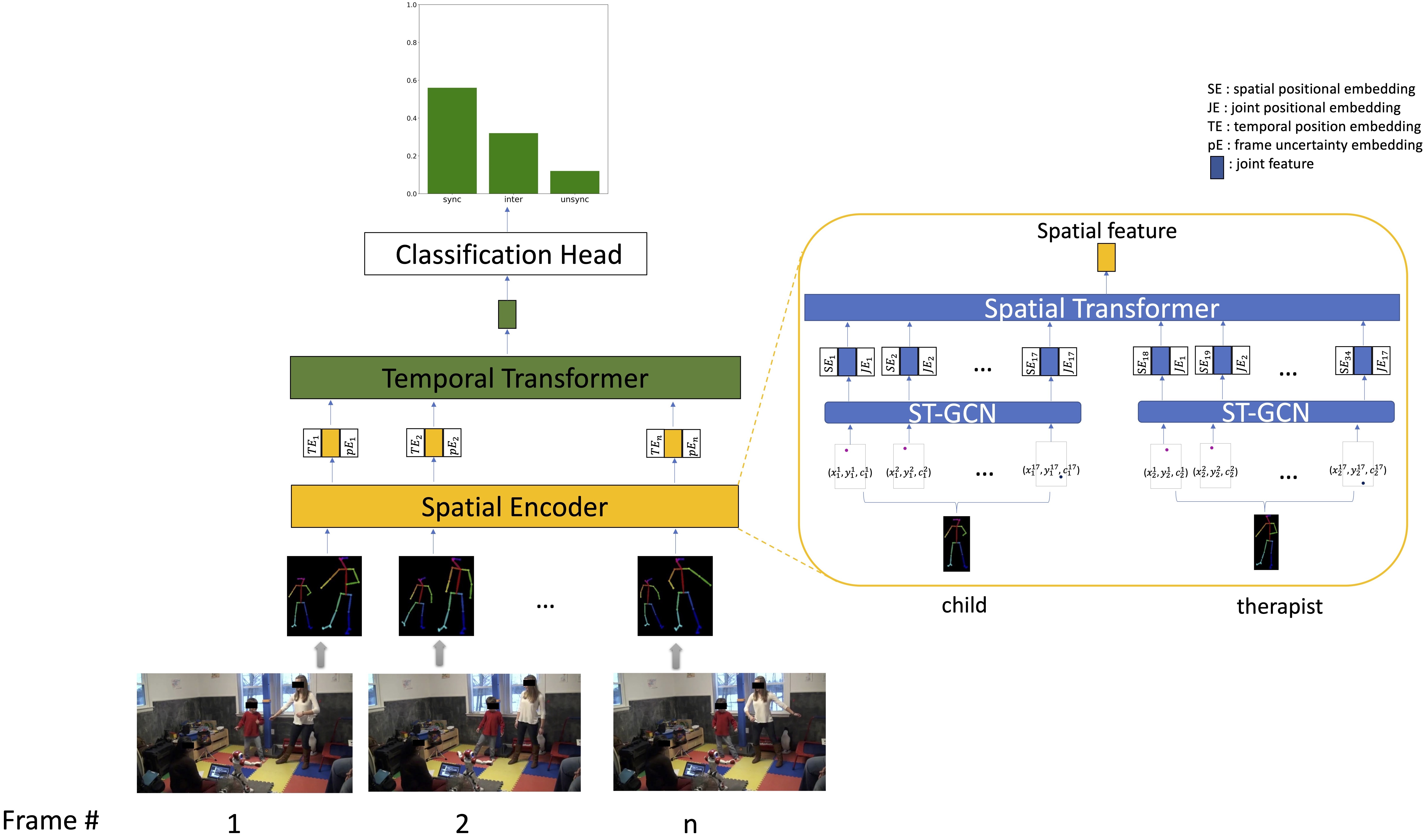}
  %Add contrastive learning and TSM in the fig?
  \caption{The proposed framework has three major components: (1) spatial encoder, (2) temporal transformer, and (3) classification head. The spatial encoder is composed of an ST-GCN \cite{yan2018spatial} and a spatial transformer.}
  \Description{Proposed framework}
  \label{fig:framework}
\end{figure}

The proposed framework is presented in Fig. \ref{fig:framework}. There are three major components: (1) a Spatial Encoder composed of ST-GCN and a spatial transformer to generate spatial features for each frame, (2) a Temporal Transformer to generate temporal features for a whole video, and (3) a Classification Head to predict synchrony class label. In this section, we first explain the major components of our framework in Section 3.1 to 3.4, then introduce two techniques to further improve the performance of our model in Section 3.5 and 3.6.  

% 1. ST-GCN
\subsection{ST-GCN}
Previous works \cite{zheng2021poseformer, li2022dyadic} generate spatial embedding for skeletons via a simple linear projection. In this work, we adopted ST-GCN \cite{yan2018spatial}, a spatial-temporal graph convolution neural network to better capture the topology among joints. 
Let $\mathcal V$ denote the vertex set on the skeleton graph where $v_i \in \mathcal V $ represents one particular body joint $J_i$, and the graph convolution operation on $\mathcal V $ can be formulated as \cite{yan2018spatial}
\begin{equation}
   \label{eq: graph_conv}
   \mathbf{f}_{out}(v_i)= \sum_{\mathcal B} \sum_{v \in \mathcal B} \frac{1}{Z_{\mathcal B}(v)} \mathbf{f}_{in}(v) \cdot \textbf{w}(l_i(v))
\end{equation}
where $f_{in}$ ($f_{out}$) denotes the feature map of $\mathcal V $  before (after) convolution. 
$\mathcal B$ is the \textit{neighbour set(s)} of $v_i$, which is determined by human body's natural connections together with a partition rule $l_i$ (i.e., distance partitioning \cite{yan2018spatial}). 
$l_i$ decides the \textit{neighbour set index} of $v$, given that there can be multiple neighbour sets.
\textbf{w} is the weighting function, and $Z$ is a normalization term that equals the cardinally of $\mathcal B$. 
In practice, Eq. \ref{eq: graph_conv} is commonly implemented as follows \cite{KipfSemi2017}: 
% to-do: what does j stands for
\begin{equation}
    \mathbf{f}_{out}=\sum_j \mathbf{\Lambda}_j^{-\frac{1}{2}} (\mathbf{A}_j \otimes \mathbf{M}_j) \mathbf{\Lambda}_j ^{-\frac{1}{2}} \mathbf{f}_{in} \mathbf{W}_j 
\end{equation}
where $j$ is the number of adjacency matrices per the partitioning rule  $l$, $\mathbf{A}_j$ is an adjacency matrix, $\mathbf{\Lambda}_j^{-\frac{1}{2}}$ is the normalized diagonal matrix of $\mathbf{A}_j$, $\mathbf{M}_j$ is a learnt attention map for each vertex (and edge), $\otimes$ denotes element-wise product, and $\mathbf{W}_j$ is the graph convolution filter determined by weighting function \textbf{w}. 

%i \in \{1, \ldots, 17\}
To preserve the uncertainty of the pose detector $D$, each joint $J_i$ is denoted as a three dimensional vector $(x_i, y_i,c_i)$, where $(x_i, y_i) \in \mathbb{R}^2$ is the coordinate of $J_i$, $c_i \in [0,1]$ is the confidence score provided by $D$ depicting the accuracy of $(x_i, y_i)$. And for each joint $J_i$, we generate a hidden embedding $\mathbf{f}_{out}(v_i)$ using ST-GCN, which is also the input of the spatial transformer as explained in Section \ref{sec:spatial_TF}. 

% 2. spatial transformer
% to-do: output dimension of spatial transformer?? 
\subsection{Spatial Transformer}
\label{sec:spatial_TF}
The spatial embedding has three components: (1) a \textit{patch embedding} $E$, (2) a trainable \textit{spatial positional embedding} $E_{pos}$, and (3) a trainable \textit{joint index embedding} $J_{pos}$ shared by the same joint within a child-therapist dyad. 
The patch embedding $E$ represents the hidden features of the input (i.e., joint coordinates) and $E_i = \mathbf{f}_{out}(v_i)$.
$E_{pos}$ retains the order of inputs which is widely used in TFNs \cite{zheng2021poseformer,dosovitskiy2020vit,Touvron2021deit,devlin2019bert,li2022dyadic}, and $J_{pos}$ is shared by the same joint between child and therapist. The motivation for introducing $J_{pos}$ is to adapt and modify TFNs to accommodate dyadic skeleton inputs, since current skeleton-based TFNs only accept one single skeleton as their input \cite{plizzari2021skeletonSTTF,liu2022graph,bai2021hierarchical}. We suggest that $J_{pos}$ can help TFNs better attend to the same joint across two persons, which is critical in deciding movement synchrony. 
The spatial feature $Z$ is generated by the addition of all three components:
\begin{equation}
     Z = \mathcal{ST}(E + E_{pos} + J_{pos})
\end{equation}
Here $\mathcal{ST}$ stands for spatial transformer which has a similar architecture as \cite{zheng2021poseformer,li2022dyadic} but has been adapted to accept dyadic skeletons.

% to-do: clarify the structure of auto-encoder
\subsection{Temporal Transformer}
The temporal embedding also has three components: (1) a \textit{patch embedding} $E' = Z$ which is the output of $\mathcal{ST}$, (2) a trainable \textit{temporal positional embedding} $E'_{pos}$, and (3) a trainable \textit{frame uncertainty embedding} $E_c$. 
$E'_{pos}$ retains the temporal frame order, and $E_c$ is computed from joint confidence within that frame. 
Specifically, for one particular frame $j$, the frame uncertainty embedding $E_c^j$ can be computed as:
\begin{equation}
    E_c^j = \operatorname{Linear} (c_1^1, \ldots, c_1^{17}, c_2^1, \ldots,c_2^{17}) 
\end{equation}
where $\operatorname{Linear}$ stands for linear projection, and $(c_1^1, \ldots, c_1^{17}, c_2^1, \ldots,c_2^{17})$ is the confidence of all joints as illustrated in Fig. \ref{fig:framework}. In addition, we also integrate a temporal self-similarity matrix $\mathcal{S}$ \cite{su2021how,dwibedi2020counting} in the computation of temporal attention, considering the periodic intrinsic of therapy interventions. $\mathcal{S}$ can be expressed by a $T \times T$ matrix where $T$ is the sequence length, and $\mathcal{S}[i][j]$ describes the similarity between \textit{pose} $X_1^i$ from the first person at time stamp $i$, and \textit{pose} $X_2^j$ from the second person at time stamp $j$. Instead of computing similarity from raw coordinates, we compute from their corresponding $d$-dimensional feature vectors $\mathsf{f}_1^i$ and  $\mathsf{f}_2^j$ generated by ST-GCN. 
The similarity function is defined as the Euclidean distance between feature vectors, followed by taking softmax over the time axis. 
Consequently, we have
\begin{equation} % 
    \mathcal{S}[i][j] =  - \frac{1}{d} \sqrt{\sum_{k=1}^{d} \| \mathbf{f}_{1}^{i}[k] - \mathbf{f}_{2}^{j}[k]} \|^{2}
    \label{eq:csm}
\end{equation}
We further apply a convolution layer on $\mathcal{S}$ to generate a feature map $\hat{S}$, and then include $\hat{S}$  in temporal attention computation \cite{su2021how}: 
\begin{equation}
    \label{eq:attn}
    \operatorname{Attention}(Q, K, V)= \psi(\frac{Q K^{T}+\hat{S}}{\sqrt{d}}) V
\end{equation}
where Q, K, V are the standard query, key and value respectively \cite{vaswani2017attention}, $\psi$ is the softmax function. An empirical \textit{scaling factor} $\sqrt{d}$ is introduced in Eq. (\ref{eq:attn}) to ensure a stable training process \cite{vaswani2017attention}.

\subsection{Classification Head}
%with Layer norm ？
Classification Head is composed of a simple $\operatorname{Linear}$ projection. Since we formulate movement synchrony estimation as a classification problem, we train our framework by Cross Entropy Loss ($\mathcal{L}_{CE}$):
\begin{equation}
    \mathcal{L}_{CE} = -\frac{1}{N} \sum_{i=1}^N \mathcal{I}(y_{i}) \cdot \log(p(\hat{y_i}))
\end{equation}
where $N$ is training data size, $y_i$ ($\hat{y_i}$) is the prediction (ground truth) for data $i$, $\mathcal{I (.)}$ is a binary indicator that outputs 0 if $y_i = \hat{y_i}$ otherwise 1, $p(\hat{y_i})$ is the predicted probability for the ground truth class $\hat{y_i}$.

\subsection{Contrastive Learning} 
%info-Nce loss; MOCO \cite{he2020moco} 
We pretrained our framework using a benchmark dataset as explained in Section \ref{sec:dataset}. 
A skeleton sequence $X$ is first augmented into two different views: a query view $X_q$ and a key view $X_k$ as explained in section \ref{sec: data_aug}. 
We adopted a contrastive learning mechanism following MoCo \cite{he2020moco}. 
MoCo builds a dynamic dictionary for contrastive learning to decouple dictionary size from mini-batch size. 
A slowly progressing key encoder $f_k$ is implemented as a momentum-based moving average of the query encoder $f_q$. Therefore, $f_k$ and $f_q$ share the same architecture but not the same weights, and \textit{only} the query encoder $f_q$ is actively updated as Equation (\ref{eq:infonce}). 
Let $E_q$ and $E_k$ be the embeddings of $X_q$ and $X_k$, namely $E_q = f_q(X_q)$, $E_k = f_k(X_k)$, we used \textit{InfoNCE loss} \cite{lai2019cpc} as contrastive loss:

% \begin{equation}
%     \label{eq:infonce}
%     \mathcal{L}_{InfoNCE}(X)=-\log \frac{\exp (Z_{q} \cdot Z_{k} / \tau)}{\exp (Z_{q} \cdot Z_{k} / \tau )+\sum_{Z_{n} \sim \mathcal{N}} \exp (Z_{q} \cdot Z_{n} / \tau)}
% \end{equation}

\begin{equation}
    \label{eq:infonce}
    \mathcal{L}_{InfoNCE}(X)=-\log \frac{\exp (E_{q} \cdot E_{k} / \tau)}{\exp (E_{q} \cdot E_{k} / \tau )+\sum_{E_{n} \sim \mathcal{N}} \exp (E_{q} \cdot E_{n} / \tau)}
\end{equation}
where $\tau$ is a temperature softening hyper-parameter \cite{lai2019cpc,thoker2021skeleton}, $N$ is the current negative sample set \cite{he2020moco} and $E_n$ is the query embedding of negative samples . 
This process helps the framework to learn transformation invariable action embeddings.

% ***Note that contrastive is on sync/inter/unsync class, not activity class; since the input of our network is two not one. still we need to assign label before this step, finished label human3.6M

% knowledge distillation
\subsection{Knowledge Distillation} 
% Cross-entropy
% Traditionally in a classification problem, our goal is to minimize the cross-entropy loss $\mathcal{L}_{\mathrm{CE}}$ between model predictions (logits) $Z_\mathrm{s}$ and ground truth labels $y$. Let $\psi$ be the softmax function
% \begin{equation}
%   \mathcal{L}_{\mathrm{CE}} = (\psi(Z_\mathrm{s}), y)
% \end{equation}

% to-do: find the slide template for neural network visualization
\begin{figure*}[h]
  \centering
  \includegraphics[width=\linewidth]{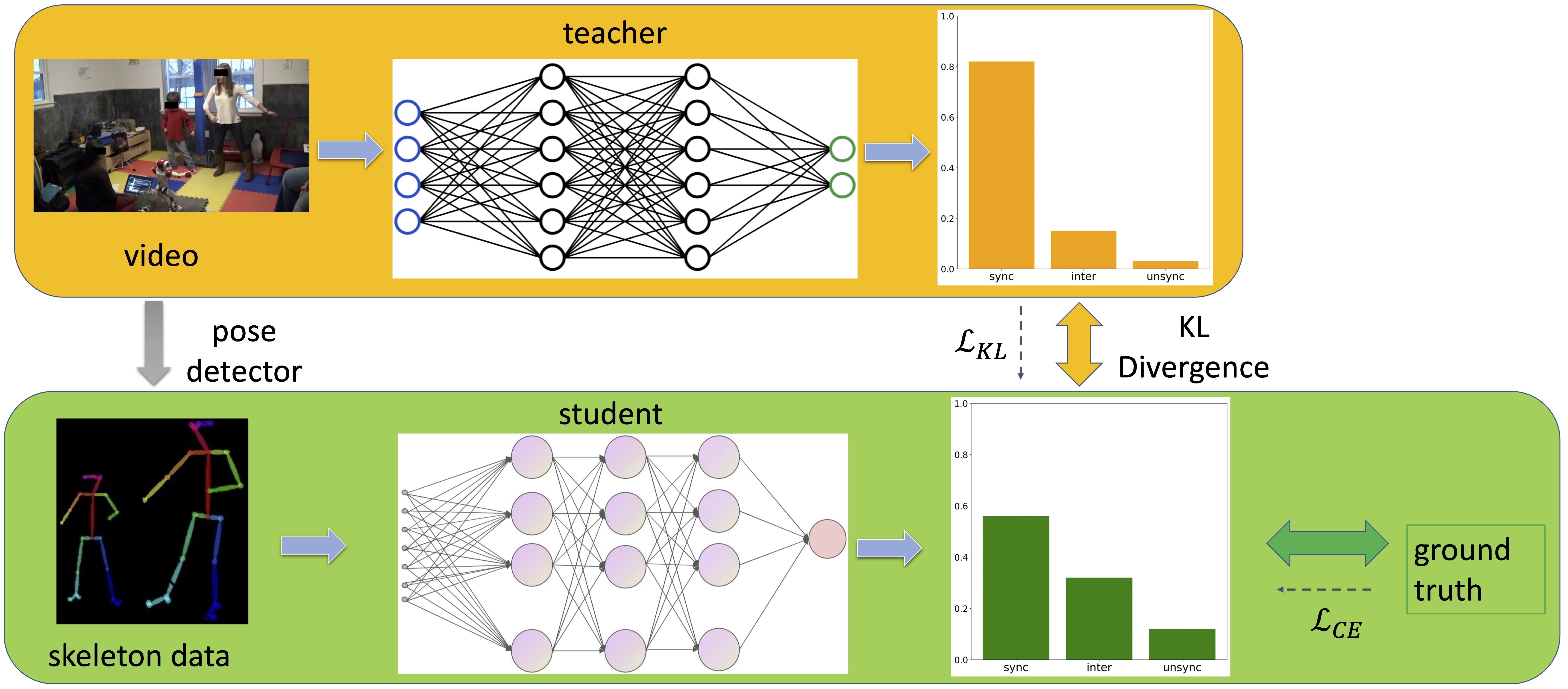}
  \caption{Knowledge distillation. The teacher model accepts original video recordings as input, while the student model accepts skeleton data derived from videos as input. The student model can learn from ground truth class labels as well as the teacher's prediction distribution (soft labels).}
  \Description{knowledge distillation}
  \label{fig:knowledge distillation}
\end{figure*}

The pipeline of knowledge distillation is visualized in Fig. \ref{fig:knowledge distillation}.
We adopted a "soft" distillation paradigm \cite{hinton2015distilling,wei2020circumventing,Touvron2021deit}, where a student model learns from the output vector of a teacher model's softmax function, and such output vector is also called \textit{soft labels}. 
% KL loss
Let $Z_\mathrm{t} (Z_{\mathrm{s}})$ be the logits of the teacher (student) model, we introduce a new regularization term $\mathcal{L}_\mathrm{KL}$ to minimize the Kullback-Leibler (KL) divergence between the predicted distribution of the teacher $p_{tea}$ and the student model $p_{stu}$
\begin{align}
       \mathcal{L}_{\mathrm{KL}} &= \tau'^{2} \cdot \mathrm{KL} (\psi(Z_{\mathrm{s}} / \tau'), \psi(Z_{\mathrm{t}} / \tau') \\
       &= -\frac{\tau'^2}{N} \sum_{i=1}^{N} \sum_{j=1}^3 p_{tea}(j) \log \frac{p_{tea}(j)}{p_{stu}(j)}
\end{align}
% predicted distribution $\hat{p}$ and ground truth distribution $p$
where $\tau '$ is the ``temperature" \cite{hinton2015distilling} for knowledge distillation. 
% combination
The overall loss for $\mathcal{L}_{KD}$ can be denoted by a weighted sum of cross entropy loss $\mathcal{L}_{\mathrm{CE}}$ and $\mathcal{L}_{\mathrm{KL}}$
\begin{equation}
    \mathcal{L}_{KD}=\alpha \mathcal{L}_{\mathrm{CE}} + (1-\alpha) \mathcal{L}_{\mathrm{KL}}
\end{equation}
where $\alpha$ is the coefficient for $\mathcal{L}_{\mathrm{CE}}$.

\section{Dataset}
\label{sec:dataset}
% to-do: add confidence score distribution
In this paper, we will first pretrained our model on Human3.6MS dataset \cite{Ionescu2014h36m} and then fine-tuned it on PT13 dataset. 

%\subsection{PT13}
\textbf{PT13} is collected from video recordings of play therapy interventions for children with autism.  It is composed of 1,273 data samples and covers 13 unique activities on various themes. 
PT13 is annotated by the level of synchrony between children and therapists within the therapeutic procedure. 
As a result, PT13 has three types of non-overlapping classes: \textit{Synchronized} (Sync), \textit{Moderately Synchronized} (ModSync) and \textit{Unsynchronized} (Unsync). 
While the Synchronized (Unsynchronized) reflects absolute (no) synchrony between children and therapists, Moderately Synchronized refers to children reacting appropriately with slight inconsistency or tardiness. 
% The bio significance is justified in Table \ref{tab: bio_significance}  between train and test splits.
% The average data length is 148 frames, with dimensions ranging from 320 by 240 pixels to 720 by 480 pixels.

% https://github.com/facebookresearch/VideoPose3D/blob/main/DATASETS.md
%\subsection{Human3.6M}
\textbf{Human3.6MS} is derived from Human3.6M \cite{Ionescu2014h36m}, one of the most popular datasets for 3D Human Pose Estimation, consists of about 3.6 million frames with 3D ground truth annotation collected by a high-speed motion capture system in an indoor setting. 
It covers 17 distinct scenarios performed by 11 professional actors, and each subject was recorded from 4 different views. Instead of using ground truth annotations, we adopted labels generated by a pose detector as provided in \cite{pavllo2019videopose3d} for consistency with PT13, especially in terms of joint uncertainty introduced by the pose detector. 
 
% to-do: add reason why choose Human3.6M
There are multiple reasons that we choose Human3.6M as the augmentation dataset for PT13. First, both datasets share similar activities, for example, ``walking" in Human3.6M and ``marching" in PT13. Activities in both datasets involve different body parts. In addition, in Human3.6M, different views of the same action provide a rich resource for ``Synchronized" data generation, since children can be treated as another ``view" of the therapist in the same scene if their actions are synchronized. The summary of both datasets is shown in Table \ref{tab:dataset}.

%test: 86+88+80 = 254
% PT13 vs. Human3.6M
\begin{table}
  \caption{Statistics of PT13 and Human3.6MS.}
  \label{tab:dataset}
  \begin{tabular}{lcc}
    \toprule
    &PT13 \cite{Li2021improving} & Human3.6MS \cite{Ionescu2014h36m}\\
    \midrule
    \# of samples  & 1,273 & 157,250\\
    sync / inter / unsync  &  432 / 441 / 400 & 50,160 / 51,330 / 55,760 \\
    train / test split & 1,019/ 254 & \textemdash \\ 
    \# of frames & 81,911 & 2,508,000\\
    \# of activities & 13  & 17 \\ 
    \# of participants &  34 & 10 \\
    joint confidence ($mean \pm std$) & $0.66 \pm 0.23$ &  $0.74 \pm 0.31$ \\
  \bottomrule
\end{tabular}
\end{table}

\subsection{Data Construction}
% to-do: visualize constructed data in supplement materials
Since there are no synchrony annotations in Human3.6M, we need to manually construct data samples in different synchrony levels to build Human3.6MS.

\textbf{Synchronized}: different views of the same action from the same actor are labeled as Synchronized. We also applied data augmentation as discussed in section \ref{sec: data_aug} to prevent the model from learning view transformations rather than critical movement features to estimate synchrony. In addition, we introduced a slight Gaussian noise to improve model robustness and mitigate domain disparity between PT13 and Human3.6M.
% a threshold $\alpha_1$

\textbf{Unsynchronized}: Unsynchronized data can be easily constructed by combining data samples from different activity classes, regardless of whether samples are from the same actor or not.    
\textbf{Moderated}: Moderated data construction is more challenging. 
We mainly generate data of this kind from: 
(1) same activities from different actors, (2) asymmetric data operation on Synchronized data. For example, one sequence is augmented as Section \ref{sec: data_aug} (i.e., joint jittering, frame shuffling/masking) while the other sequence remains unchanged, or some frames in one sequence are randomly replaced by data from a different activity.

All constructed data samples can be further augmented following Section \ref{sec: data_aug}.

\subsection{Synchrony Invariant Skeleton Augmentation}
\label{sec: data_aug}
In this section, we introduce synchrony invariant skeleton data augmentation methods applied in our work as an extension of \cite{thoker2021skeleton}. 
These methods fall into two categories, namely spatial augmentation and temporal augmentation. 
Spatial augmentation manipulates joint coordinates, while temporal augmentation transforms the temporal order across the frames. 
The overall augmentation is a combination of all kinds and is identically applied to both skeletons of a pair.
% rewrite the following sentence
Therefore, for simplicity, we only explain the operation on one skeleton sequence instead of two. 
Let $(T,X,Y,C)$ be a skeleton sequence $S$, $(X,Y,C)$ denote a pose $p \in S$ at one particular frame, and $(x,y,c)$ denote a joint $j \in p$. Augmentation on $p$ will be applied identically to all joints $j \in p$, while augmentation on a joint $j$ only change its own coordinates $(x,y,c)$.

\subsubsection{Spatial Skeleton Augmentation}
\begin{enumerate}
    \item shifting: apply a translation of $(\Delta x, \Delta y)$ on a joint $j$. The new coordinate is $(x+ \Delta x, y+ \Delta y,c)$.
    \item flipping: apply left-right flipping on joint coordinates. 
    \item shearing \cite{thoker2021skeleton}: apply the same shearing operation to each joint of $p$ to simulate changes in camera viewpoint and distance between the subject and camera.
    \item jittering: a subset of the joint connections are randomly perturbed, namely the selected joints are moved to irregular positions while all the others remain unchanged. 
    \item spatial masking: a small subset of $N (N \leq 3)$ joints are masked out, and their coordinates $(x,y,c)$ are adjusted to $(0,0,0)$.
\end{enumerate}

\subsubsection{Temporal Skeleton Augmentation}
\begin{enumerate}
    \item shuffling: randomly shuffle the temporal order of $S$.
    \item repeating: replace a subset of frames in $S$ by repeating their temporal neighbours. Moderated data is \textit{excluded} from this operation.
    \item temporal masking: for a small subset of frames in $S$, assign $p = (X,Y,C)$ to $(\textbf{0}, \textbf{0}, \textbf{0})$.
\end{enumerate}
% frame shuffling: pro: most likely to keep the original sync level. Cons: potentially destroy semantic info; disentangled intervention activity 

% to-do: visualize data augmentation
% \begin{figure}[h]
%   \centering
%   \includegraphics[width=\linewidth]{paper/figures/augmentation.jpg}
%   \caption{left: spatial augmentation; right: temporal augmentation}
%   \Description{data augmentation}
% \end{figure}

\section{Experiments}
\label{sec:experiment}
\subsection{Implementation Details}
We implemented our proposed framework with Pytorch \cite{paszke2019pytorch}. 
Eight NVIDIA Tesla V100 GPUs were used for training and testing.
We adopted the pretrained ST-GCN on Kinetics \cite{kay2017kinetics} and froze its weights during training. 
The hyperparameters of spatial / temporal transformer is similar to \cite{li2022dyadic}.
We chose a sequence length $T = 81$ and adopted COCO \cite{lin2014coco} format representation for skeleton, where each pose is composed of 17 keypoints.
All hidden embeddings in the spatial/temporal transformer are set to the same dimension $d   = 544$ due to the residual connections within the transformer.
The temperature $\tau$ in Eq. (\ref{eq:infonce}) is set as 0.07 \cite{he2020moco}.
We take the typical values  $\alpha = 0.9$ and $\tau' = 3.0$ for knowledge distillation \cite{Touvron2021deit}.
We pretrained our model on Human3.6MS for 130 epochs and fine-tuned it on PT13 for 800 epochs using Adam optimizer \cite{Diederik2015adam} with a batch size of 64. 
We adopted an exponential learning rate decay scheduler with the initial learning rate of 1e-3 and a decay factor of 0.98 for each epoch.
The dropout \cite{srivastava2014dropout} rate was set to $0.5$ in the training phase to prevent overfitting. 

\subsection{Synchrony Classification on PT13}
We compared our method with representative counterpart approaches  \cite{berndt1994dtw,Pedregosa2011scikit-learn,coco2014cross, gao2020asymmetric, li2022dyadic} for movement synchrony estimation on PT13 as shown in Table \ref{tab: performance}, including both statistical and deep learning approaches. 
Specifically, statistical approaches \cite{berndt1994dtw,Pedregosa2011scikit-learn,coco2014cross} used support vector machine \cite{cortes1995svm} as classification head.
Overall, our method outperforms its counterparts by a considerable margin in classification accuracy, especially for the Moderately Synchronized class.

% to-do: fix F1 -score
We also reported a row-wisely normalized confusion matrix of our model on PT13 in Table \ref{tab:cm_pt13} with a F1-score of 0.89. 
The confusion matrix reflects a robust performance of our model across all synchrony classes.

% \begin{table*}[h]
% \caption{Performance comparison on PT13 reported by classification accuracy $(\%)$.}
% \label{tab: performance}
% \centering
% %\begin{tabular}{|c|c|c|c|c|}
% \begin{tabular}{lcccc}
%     \toprule
%      & Sync & ModSync & Unsync & Avg.\\
%     \midrule
%     % baseline method
%     DTW \cite{berndt1994dtw} & 79.31& 76.40 & \textbf{87.50} &80.86\\
%     %\hline
%     2D Correlation \cite{Pedregosa2011scikit-learn} & 48.28 & 37.08 & 22.50 & 36.33\\
%     %\hline
%     % cross-wavelet transform\cite{grinsted2004cwt} & & & & & & &\\
%     % \hline
%     Cross-recurrence \cite{coco2014cross} & 86.21 & 71.91 & 16.25 & 59.38\\
%     % \hline
%     AIM \cite{gao2020asymmetric} & 62.58 & 35.66 & 40.25 & 46.14\\
%     %\hline
%     Li \etal \cite{li2022dyadic} &89.66 &80.90 & 85.00 & 85.16\\
%     \hline
%     % our method
%     Ours &\textbf{94.25} & \textbf{85.39} & \textbf{87.50} &\textbf{89.06}  \\
%     \bottomrule
% \end{tabular}
% \end{table*}

\begin{table*}[h]
\caption{Performance comparison on PT13 reported by classification accuracy $(\%)$.}
\label{tab: performance}
\centering
%\begin{tabular}{|c|c|c|c|c|}
\begin{tabular}{lcccc}
    \toprule
     & Sync & ModSync & Unsync & Avg.\\
    \midrule
    % baseline method
    DTW \cite{berndt1994dtw} & 79.07& 76.14 & \textbf{87.50} &80.71\\
    %\hline
    2D Correlation \cite{Pedregosa2011scikit-learn} & 48.84 & 37.50 & 22.50 & 36.61\\
    %\hline
    % cross-wavelet transform\cite{grinsted2004cwt} & & & & & & &\\
    % \hline
    Cross-recurrence \cite{coco2014cross} & 86.05 & 71.59 & 16.25 & 59.06\\
    % \hline
    AIM \cite{gao2020asymmetric} & 62.79 & 35.23 & 40.00 & 46.06\\
    %\hline
    Li \etal \cite{li2022dyadic} &89.53 &80.68 & 85.00 & 85.04\\
    \hline
    % our method
    Ours &\textbf{94.19} & \textbf{85.23} & \textbf{87.50} &\textbf{88.98}  \\
    \bottomrule
\end{tabular}
\end{table*}

% array([[97.70114943,  1.14942529,  1.14942529],
%       [12.35955056, 85.39325843,  2.24719101],
%       [ 5.        ,  2.5       , 92.5       ]])

\begin{table}[h]
    \caption{Normalized confusion matrix on PT13 ($\%$).}
    \label{tab:cm_pt13}
    \centering
    \begin{tabular}{l|l|c|c|c|c|}
    \multicolumn{2}{c}{}&\multicolumn{2}{c}{Prediction}\\
    \cline{3-5}
    \multicolumn{2}{c|}{} & Sync & ModSync & Unsync \\
    \cline{2-5}
    \multirow{2}{*}{Label}& Sync & 94.19&3.49&2.32\\
    \cline{2-5}
    &ModSync & 12.34&85.23& 2.43\\
    \cline{2-5}
   &Unsync & 5.00 & 7.50 & 87.50\\
    \cline{2-5}
    \end{tabular}
\end{table}

\subsection{Ablation Study}
To verify the contribution of each component in our framework, we conducted extensive ablation experiments as follows.

\subsubsection{ST-GCN}
ST-GCN considers joint topology in the generation of spatial embeddings for a skeleton. In order to justify its advantage over a simple linear projection layer as applied in previous works \cite{li2022dyadic, zheng2021poseformer}, we compared their performance in Table \ref{tab: ablation_gcn}. Results show that ST-GCN outperforms a linear projection layer across all three synchrony classes.

\begin{table*}[h]
\caption{Ablation study: ST-GCN. Reported by classification accuracy on PT13 ($\%$).}
\label{tab: ablation_gcn}
\centering
%\begin{tabular}{|c|c|c|c|c|}
\begin{tabular}{ccccc}
    \toprule
     & Sync & ModSync & Unsync & Avg.\\
    \midrule
    Linear projection & 90.68 & 81.81 & 85.00 & 85.83\\
    % our method
    Ours &\textbf{94.19} & \textbf{85.23} & \textbf{87.50} &\textbf{88.98} \\
    \bottomrule
\end{tabular}
\end{table*}

\subsubsection{Temporal Similarity Matrix}
In order to evaluate that incorporating a temporal similarity matrix can better capture periodic movement features, we conducted experiments under three different settings in temporal attention computation: (1) exclude TSM, (2) include TSM computed from raw coordinates, (3) include TSM computed from feature vectors produced by the spatial encoder.
Table \ref{tab: ablation_tsm} demonstrates the benefits of TSM in our model.

\begin{table*}[h]
\caption{Ablation study: Temporal Similarity Matrix. Reported by classification accuracy on PT13  ($\%$).}
\label{tab: ablation_tsm}
\centering
%\begin{tabular}{|c|c|c|c|c|}
\begin{tabular}{ccccc}
    \toprule
     & Sync & ModSync & Unsync & Avg.\\
    \midrule
    % baseline method
    Without TSM & 89.53 & 82.95 & 86.25 &86.22\\
    TSM from coordinates &90.70 & 84.09& \textbf{87.50} & 87.40\\
    % our method
    Ours &\textbf{94.19} & \textbf{85.23} & \textbf{87.50} &\textbf{88.98}\\
    \bottomrule
\end{tabular}
\end{table*}

\subsubsection{Joint Uncertainty} 
Table \ref{tab:dataset} reveals that joint uncertainty commonly resides as suggested by the mean and variance of joint confidence.
In this paper, we introduced joint uncertainty both in spatial and temporal feature generation. 
We defined two baseline models: (1) plain baseline: exclude spatial uncertainty by assigning all joint confidence to 1 for input format consistency, exclude frame uncertainty embedding; 
and (2) spatial baseline: include joint confidence in spatial feature generation only, exclude frame uncertainty embedding.
We compared these two baselines with our model in Table \ref{tab:ablation_uncertainty}, and our model achieved a better performance in movement synchrony estimation compared to provided baseline methods. 
This could be explained by our model attending to joints (frames) where there were high confidence and thus producing more convincing results. 
% One reasonable explanation to the superior performance of our model compared to baselines is that joint (frame) confidence guided our model to attend to joints (frames) of high confidence in movement synchrony estimation.

\begin{table*}[h]
\caption{Ablation study: Joint Uncertainty. Reported by classification accuracy on PT13  ($\%$).}
\label{tab:ablation_uncertainty}
\centering
%\begin{tabular}{|c|c|c|c|c|}
\begin{tabular}{ccccc}
    \toprule
     & Sync & ModSync & Unsync & Avg.\\
    \midrule
    % baseline method
    Plain baseline & 87.21 & 80.68 & 86.25 &84.65\\
    Spatial baseline &88.37 & 81.81& 86.25 & 85.43\\
    % our method
    Ours &\textbf{94.19} & \textbf{85.23} & \textbf{87.50} &\textbf{88.98}\\
    \bottomrule
\end{tabular}
\end{table*}

\subsubsection{Joint Positional Embedding}
In addition to traditional positional embeddings widely applied in TFNs \cite{zheng2021poseformer,dosovitskiy2020vit,Touvron2021deit,devlin2019bert,plizzari2021skeletonSTTF} to maintain patch order, in this work we took a novel approach by introducing a joint positional embedding $JE$ shared between the same joint of interacting dyads.
% $JE$ provides extra guidance to our network to attend to the same joint across different subjects in movement synchrony estimation.
Our network is provided with additional guidance by $JE$ to focus on the same joint across different subjects in movement synchrony estimation, resulting in a positive change in performance (see Table \ref{tab: shared_joint_embed}).
It is also one of the major modifications to the spatial transformer to adapt to dyad skeletons, while most TFNs \cite{liu2022graph, bai2021hierarchical, zheng2021poseformer} only take one single skeleton as input.   

\begin{table*}[h]
\caption{Ablation study: Joint Positional Embedding. Reported by classification accuracy on PT13  ($\%$).}
\label{tab: shared_joint_embed}
\centering
%\begin{tabular}{|c|c|c|c|c|}
\begin{tabular}{ccccc}
    \toprule
     & Sync & ModSync & Unsync & Avg.\\
    \midrule
    Without shared joint embedding & 90.69 & 82.95 & 86.25 & 86.61\\
    % our method
     Ours &\textbf{94.19} & \textbf{85.23} & \textbf{87.50} &\textbf{88.98}\\
    \bottomrule
\end{tabular}
\end{table*}

\subsection{Contrastive Learning}
Inadequately pretrained TFNs can have worse performance than counterpart methods with simpler architectures.
We justify the importance of proper pretraining by comparing the performance of our model with and without contrastive learning using Human3.6MS.
As Table \ref{tab:contrastive} shows, the great margin supports the significance of pretraining.

\begin{table*}[h]
\caption{Comparison of classification accuracy with/out contrastive learning on PT13 ($\%$).}
\label{tab:contrastive}
\centering
%\begin{tabular}{|c|c|c|c|c|}
\begin{tabular}{ccccc}
    \toprule
     & Sync & ModSync & Unsync & Avg.\\
    \midrule
    Ours (w/o pretraining) & 84.88 & 78.41 & 80.00 & 81.10\\
    % our method
    Ours (with pretraining) &\textbf{94.19} & \textbf{85.23} & \textbf{87.50} &\textbf{88.98}  \\
    \bottomrule
\end{tabular}
\end{table*}

\subsection{Knowledge Distillation}
This section demonstrates how knowledge distillation can alleviate information loss introduced by pose detector failure in a privacy-preserving manner. 
Here, we define \textit{privacy} as having no access to raw video.
We adapted the standard teacher-student diagram:
a teacher model was trained on original video recordings by therapists, while a  student model (our model) was trained on skeleton data derived from original videos. 
Table \ref{tab: knowledge_ditillation} shows a considerable accuracy improvement of our model after the application of knowledge distillation. 
The end-to-end design of the teacher model ensures that therapists who possess an essential machine learning background can effectively train the model.

\begin{table*}[h]
\caption{Comparison of classification accuracy before/after knowledge distillation on PT13 ($\%$).}
\label{tab: knowledge_ditillation}
\centering
%\begin{tabular}{|c|c|c|c|c|}
\begin{tabular}{cllll}
    \toprule
     & Sync & ModSync & Unsync & Avg.\\
    \midrule
    % baseline method
    Teacher model \cite{Li2021improving} & 97.67 & 88.64 & 92.50 & 92.91\\
     \midrule
    Ours &94.19 & 85.23 & 87.50 &88.98 \\
    % our method
    Ours + knowledge distillation & 95.34 $  (\textbf{1.15 \%} \uparrow)$ & 87.50 $ (\textbf{2.27 \%} \uparrow)$ & 90.00 $ (\textbf{2.50 \%} \uparrow)$ & 90.94 $ (\textbf{1.96 \%} \uparrow)$  \\
    \bottomrule
\end{tabular}
\end{table*}

% to-do: add experiment analysis
\section{Discussion}
\label{sec:discussion}
%1.definition of privacy, especially in knowledge distillation,
To clarify, in this work, privacy refers to not accessing data that can expose human identities, such as video recordings, images  and voices, which is crucial in autism research.
The teacher model we adopted in knowledge distillation is not privacy-preserving since it requires access to original video recordings; therefore, the training process of the teacher model was conducted by therapists with fundamental machine learning backgrounds. 
%2. why need skeleton based method if video based have better performance: data sharing barrier, privacy preserving, also unfair comparison between video based and skeleton model directly
The privacy concern also explains why a skeleton-based approach for movement synchrony estimation is still essential, even though there exists a method with better accuracy but can lead to privacy violations. 
%  a simple replacement of distribution learning
In addition, the process of knowledge distillation can be potentially replaced by label distribution learning \cite{geng2016label}, and the label distribution (comparable to ``soft labels" in knowledge distillation) can be obtained by introducing more annotators.
However, extra annotation is more costly and time-consuming, another common roadblock in autism research. 
% 4. limitation of our work
% better augmentation strategy
One major limitation of our work is synchrony Human3.6MS construction, where each Synchronized / Moderately Synchronized sample was  generated (partially) from the same subjects.
This is owing to the shortage of dataset with synchrony labels, and it remains an open challenge in this work.
In addition, the fundamental idea behind data construction is to treat one person (child) as an alternate ``view" of the other person (therapist), which may not perfectly hold in PT13 as opposed to original Human3.6M.

\section{Conclusion}
\label{sec:conclusion}
% Deep learning applications in movement synchrony estimation has just caused attention. 
Deep learning applications in movement synchrony estimation have recently attracted the attention of researchers.
Considering the benefits of skeleton data in preserving critical motion features and privacy, in this paper, we proposed a skeleton-based graph transformer for movement synchrony estimation while addressing pose uncertainty.
To accommodate a pair of skeletons from interacting dyads, we uniquely designed  shareable joint embeddings and incorporated a temporal similarity matrix in attention computation.
We pretrained our model on a constructed dataset from Human3.6M, a large-scale benchmark dataset for human activity recognition.
We also provided several synchrony-invariant methods for skeleton data augmentation. 
Our work has implications for overcoming the data shortage barrier in autism research by carefully leveraging publicly accessible benchmark datasets, especially for activity comprehension in therapy intervention.
Finally, to mitigate information loss from skeleton data, we innovatively embedded joint uncertainty into our model and guided our model by soft labels derived from a teacher model, which is trained on original video recordings by professionals and therapists. 

% future work.
In future work, we plan to consider the frequency and difficulty of an action in synchrony estimation, which we did not emphasize adequately in this paper.
We will also investigate other privacy-preserving data modalities, such as gaze data \cite{guo2021icmi,Guo_Barmaki_2020}, facial landmarks, and emotion \cite{li21twostage}. 
Other group-based interaction evaluation approaches such as group activity recognition \cite{gavrilyuk2020actor} are also worth investigating.

% \begin{acks}
% \end{acks}

\bibliographystyle{ACM-Reference-Format}
\bibliography{sample-base}


\end{document}